\documentclass[lettersize,journal]{IEEEtran}
\usepackage{amsmath,amsfonts}
\usepackage{algorithmic}
\usepackage{algorithm}
\usepackage{array}
\usepackage[caption=false,font=normalsize,labelfont=sf,textfont=sf]{subfig}
\usepackage{textcomp}
\usepackage{stfloats}
\usepackage{url}
\usepackage{verbatim}
\usepackage{graphicx}
\usepackage{cite}
\usepackage{threeparttable}

\makeatletter
\renewcommand{\maketag@@@}[1]{\hbox{\m@th\normalsize\normalfont#1}}%
\makeatother

\usepackage{tikz,xcolor,hyperref}

\definecolor{lime}{HTML}{A6CE39}
\DeclareRobustCommand{\orcidicon}{
\begin{tikzpicture}
\draw[lime, fill=lime] (0,0)
circle[radius=0.16]
node[white]{{\fontfamily{qag}\selectfont \tiny \.{I}D}}; 
\end{tikzpicture}
\hspace{-2mm}
}
\foreach \x in {A, ..., Z}{%
\expandafter\xdef\csname orcid\x\endcsname{\noexpand\href{https://orcid.org/\csname orcidauthor\x\endcsname}{\noexpand\orcidicon}}
}

\begin{document}

\title{A Multi-modal Deformable Land-air Robot for Complex Environments}

\author{Xinyu Zhang\hspace{-1.5mm}\orcidA{}, Yuanhao Huang\hspace{-1.5mm}\orcidB{}, Kangyao Huang\hspace{-1.5mm}\orcidC{}, Xiaoyu Wang, Dafeng Jin, Huaping Liu\hspace{-1.5mm}\orcidD{} and Jun Li

\thanks{This work was supported by the National High Technology Research and Development Program of China under Grant No. 2018YFE0204300 and the National Natural Science Foundation of China under Grants No. 62273198 and U1964203. We thank LetPub (www.letpub.com) for linguistic assistance and pre-submission expert review.}

\thanks{Xinyu Zhang, Xiaoyu Wang, Jun Li are with The School of Vehicle and Mobility, Tsinghua University, Beijing, P.R.China.
        {\tt\small xyzhang@tsinghua.edu.cn}}%
\thanks{Yuanhao Huang is with The School of Aviation, Inner Mongolia University of Technology, Hohhot, Inner Mongolia, P.R.China
        {\tt\small huangyuanhao\_work@163.com}}%
\thanks{Kangyao Huang and Huaping Liu are with The Department of Computer Science and Technology, Tsinghua University, Beijing, P.R.China. Kangyao Huang is the corresponding author.
        {\tt\small huangky22@mails.tsinghua.edu.cn}}%
\thanks{Dafeng Jin is with Suzhou Automobile Research Institute, Tsinghua University, Suzhou, P.R.China.
        {\tt\small jindf@tsinghua.edu.cn}}%
}

\markboth{Journal of \LaTeX\ Class Files,~Vol.~14, No.~8, August~2021}%
{Shell \MakeLowercase{\textit{et al.}}: A Sample Article Using IEEEtran.cls for IEEE Journals}


\maketitle

\begin{abstract}

Single locomotion robots often struggle to adapt in highly variable or uncertain environments, especially in emergencies. In this paper, a multi-modal deformable robot is introduced that can both fly and drive. Compatibility issues with multi-modal locomotive fusion for this hybrid land-air robot are solved using proposed design conceptions, including power settings, energy selection, and designs of deformable structure. The robot can also automatically transform between land and air modes during 3D planning and tracking. Meanwhile, we proposed a algorithms for evaluation the performance of land-air robots. A series of comparisons and experiments were conducted to demonstrate the robustness and reliability of the proposed structure in complex field environments.

\end{abstract}


\section{INTRODUCTION}

\IEEEPARstart{F}{ield} tasks such as post-disaster mapping, situational awareness and analysis, search and rescue, and even special operations typically require robots that can move and work in uncertain and complex environments. However, it is often difficult for conventional single locomotion robots to adapt in these conditions. In contrast, multi-modal robots exhibit much higher flexibility and adaptability by combining wheels, legs, propellers, wings, fins, and hybrid actuators to achieve motion\cite{Rafeeq2021}. For example, land-air robots \cite{qin2020hybrid,meiri2019flying,Kalantari2013a,Dudley2015a,zhang2022autonomous} offer both flying and driving capabilities for increased spatial flexibility while reducing the energy demands of continuous flight.

Similarly, hybrid vehicles with rotary wings have shown potential for seamless and dynamic switching between modalities \cite{zhang2022intelligent}. Many of these systems are comprised of a quad-rotor hinged at the center of a cylindrical cage or a pair of coaxial passive wheels. While this design offers several advantages for reduced energy consumption, increased duration, and expanded coverage, dust caused by rotor downwash effects may reduce perception capabilities while driving \cite{wen2019numerical}. Another class of land-air robots offers both active aerial and terrestrial motion \cite{Ratsamee2016a,Kalantari2020a}. Active wheels are required in these systems to enhance ground moving performance \cite{tan2021multimodal,Sharif2019c,kalantari2020drivocopter}.


The field of small land-air robotics has seen significant progress in recent years, due to highly integrated electronic components \cite{james2021high}. Although these robots balance athletic ability and agility, their small size limits them to short-term reconnaissance missions. In contrast, larger land-air robots can increase the duration of an operation and can carry heavier loads, but their large size can be difficult to maneuver in certain terrains\cite{tan2021multimodal}. Therefore, a land-air robots are difficult to have both strong flight ability and ground mobility. Previous studies have introduced deformable mechanisms to fold propeller to change size in the design of drones and land-air robot. However, most of these designs are based on small platforms \cite{falanga2018foldable,Hu2018a}, since large deformations can limit actuator performance and restrict motion. 

Land-air robots' performance, especially ground-air movement capabilities, should be considered and compared comprehensively. Rather than just comparing single performance parameters such as energy efficiency, hovering time, or payload capacity. The main parameters of the comprehensive motion performance of land and air robots include hovering time, payload, cruise flying speed, driving mileage, obstacle height, cruise driving speed, weight, and projection in two motion modes. As such, this study aims to develop a novel land-air robot with excellent ability to fly and drive. The proposed overall design conception in Section II can solve compatibility issues for multi-modal locomotive fusion with an hybrid land-air robot generally. Automatic deformable arms with fixed propellers offer stronger flight thrust than the same size robots. On this basis, the robot can become compact and move flexibly after folding its arms, and it is also the first land-air robot with automatic folding arms. According to the motive and passable experiments with rich scenarios and quantitative performance comparisons with related robots, we verify the robot's capabilities of flying and driving. The primary technical contributions of this study can be summarized as follows: 

\begin{itemize}
    \item Proposed a generic mechtronics design conception of land-air robot, considering the power settings, energy selection, lightweight design and deformable structure. The conception ensures the land-air robot's comprehensive movement ability, including flying, driving, and modes switching.
    \item A land-air robot with automatic deformable arms is proposed for the first time. Combined with the introduced design conceptions, solving the land-air robots' compatibility issues between flying and driving modes. Larger propellers can generate greater thrust while reducing the plane projection area by 79 percent, which significantly improved passing capabilities.
    \item The proposed robot and mechanisms were verified in various simulated and practical field scenarios. The experiments and quantitative performance evaluation method provided strong evidences for their effectiveness and robustness.
\end{itemize}

\begin{figure*}[t]
\centering
\includegraphics[width=15cm]{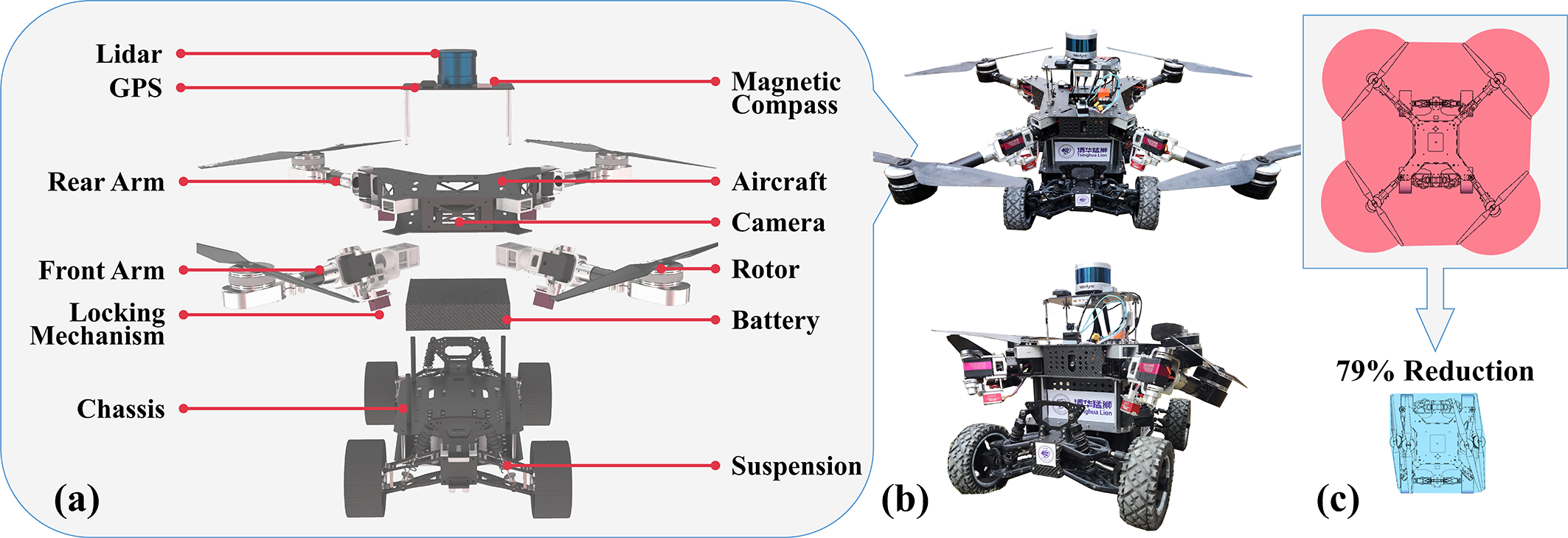}%
\caption{The overall structure of the proposed robot, including the physical display and deformation advantages. (a) Key components, including flight, ground travel, and hardware elements. (b) Flight and driving modes. (c) The vertical projection area of the robot is reduced by 79 percent compared with the unfolded state.}
\label{structure}
\end{figure*}

The remainder of this paper is organized as follows. Section II introduces the design concept and corresponding analysis. Section III describes the detailed structure and deformable mechanism controls. Section IV demonstrates the robot's overall motion and control strategy, which mapped its logic to the mechatronics system, clearly. Section V provides the results of experiments and Section VI concludes the paper. A video documenting experiments conducted as part of the study is available at \href{https://youtu.be/kGzKuHy6CmA}{video}.

\section{Design Concepts and Analysis}

The design of the multi-modal land-air robot is not a simple combination, as multiple components (e.g., power and electronic systems) are shared to reduce the total weight to the extent possible. The compatibility problems of the motion speed characteristic requirements and controls present significant challenges \cite{xie2021adaptive}. In this section, we propose design metrics and concepts for power settings, energy selection, and the deformable structure of the land-air robot. The overall design and key parameters are shown in Fig. \ref{structure} and Table \ref{specification}.

\subsection{Overall Design}

The robot configuration depends heavily on the requirements of specific tasks and corresponding environments. The various scenes, such as the doors, passages, outdoor jungle scenes, et al., clearly limit the robot's width. Thus, the robot's dimensions in driving mode to be as low as possible to allow for flexible motion and passage through narrow spaces for rapid movement. Despite these size restrictions, the robot is still expected to carry heavy equipment, including sensing devices, computing units, and actuators for improved situational awareness. Taking these configurations and load capacity demands into consideration, the maximum power and take-off weight can be estimated using existing techniques \cite{Shi2017,dai2019analytical}. Hereafter, we refer to the multi-copter flight evaluation strategy to optimize our approach, which provides a complete design and selection methodology for estimating and maximizing flight duration \cite{He2019}.

\begin{table}[h]
\renewcommand\arraystretch{1.1}
\begin{center}
\caption{\textbf{Robot Specifications}}
\label{specification}
\vspace{-0.5 mm}
\begin{tabular}{c c c}
\hline \text { Width } & \text { Expanding } & 1250 {~mm} \\
& \text { Folding } & 585 {~mm} \\
\hline \text { Depth } & \text { Expanding } & 1250 {~mm} \\
& \text { Folding } & 670 {~mm} \\
\hline \text { Height } & \text { Body } & 552 {~mm} \\
\hline \text { Weight } & \text { Total } & 24.62 {~kg} \\
& \text { Battery } & 5.74 {~kg} \\
& \text { Flight Module } & 7.61 {~kg} \\
& \text { Chassis } & 7.07 {~kg} \\
& \text { Payload } & 4.20 {~kg} \\
\hline \text { Speed } & \text { Max Flying Speed } & 7.47 {~m}/{s} \\
& \text { Max Driving Speed } & 10.06 {~m}/{s} \\
\hline \text { Flight Module } & \text { Flight Motor } & 5 {Nm} \\
& \text { Kv Value } & 170 \\
& \text { Propeller } & 26 {inch} \\
& \text { Steering Gear} & 8 {Nm} \\
\hline \text { Chassis } & \text { Kv Value } & 800 \\
& \text { Driving Motor } & 12 {Nm} \\
\hline \text { Battery } & \text { Lithium Battery } & 48{~V}-30 {Ah} \\
& \text { Duration } & 21.6 {~min} \text { Flight Time } \\
\hline
\end{tabular}
\end{center}
\end{table}

\subsection{Flight Duration Analysis}

\begin{figure*}[t]
\centering
\includegraphics[width=17.5cm]{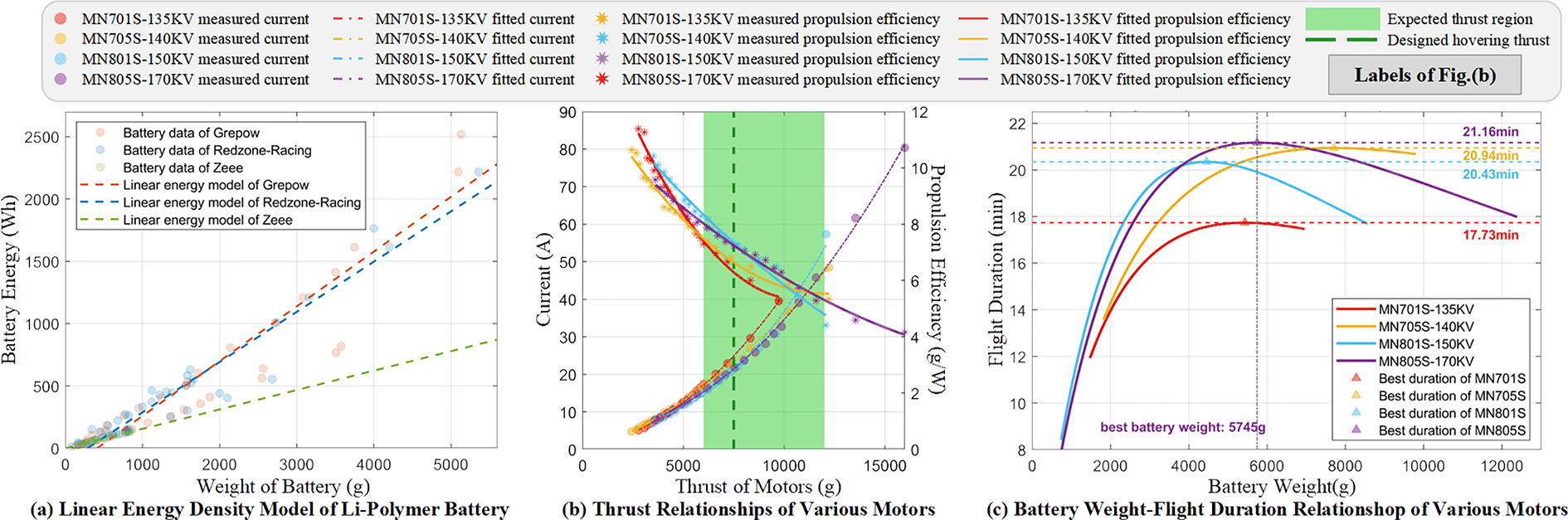}%
\caption{Optimal flight duration analysis and power.}
\label{flight duration modeling}
\vspace{-4 mm}
\end{figure*}

\begin{figure}[t]
    \centering
    \includegraphics[width=6.5cm]{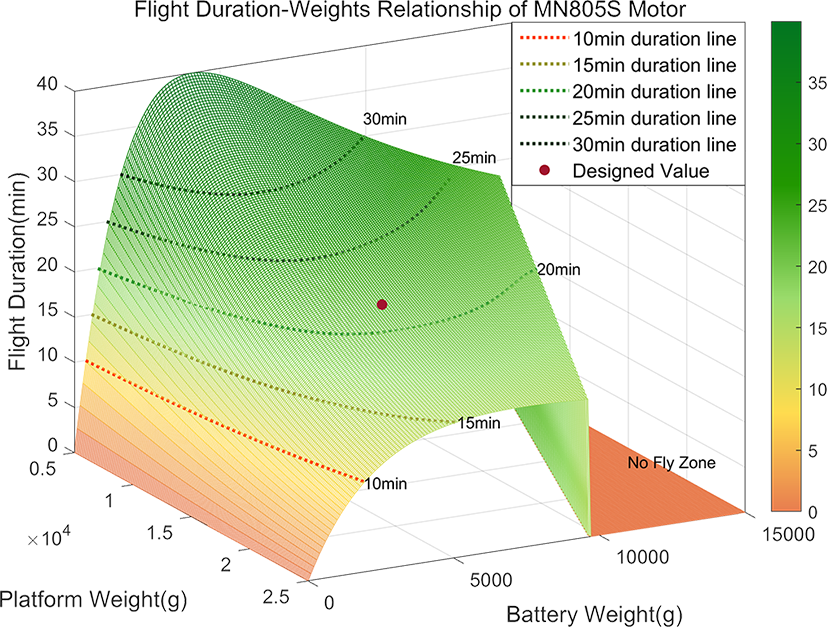}
    \caption{Optimal flight duration for various robot weight configurations.}
    \label{duration weight relationships}
     \vspace{-5 mm} 
\end{figure}

Due to redundant structures and actuators' weight, the trade-off between battery weight and flight duration is one of the most challenging aspects of the land-air robot design process. We first evaluated flight duration using battery capacity and electrical current modeling for the motor/propeller system. We compared three different brands of batteries and obtained linear models using data fitting, as shown in Fig. \ref{flight duration modeling}(a). The maximum propeller size is fixed as 26 inch, we tentatively selected four power train configurations for use as assessment samples, with brushless motors (T-Motor) varying from 135$\sim$170 KV. Corresponding data combinations were then fitted into mappings, including variations in measured current and propulsion efficiency with thrust. As shown in Fig. \ref{flight duration modeling}(b), the MN805S-170KV model can provide a large thrust range covering the expected region, with a relatively high propulsion efficiency that facilitates a higher energy efficiency for the entire power train.

The flight duration $t$ can be estimated as follows:
\begin{equation}
    t=\frac{wm}{\kappa VI+P}, P=P_{avionics}+P_{perception}+P_{deform},
    \label{flight duration}
\end{equation}
where $w$ and $m$ denote the energy density and mass of the battery, respectively, $\kappa$ represents the number of motors in the multi-copter (a constant), $V$ is the nominal motor voltage, $I$ is the instantaneous current, and $P$ is the total power, consisting of power consumed by avionics systems, sensing devices, and deformations. Onboard power consumption can be calculated from the rated power and the mass of the robot. We set the calculated power rating to the output power when the robot hovering and was treated as a constant. Hovering thrust generated by the motors at rated power should be greater than or equal to the robot's weight. Thus, we can assume the hovering thrust is approximately equal to the combined weight of these modules:
\begin{equation}
    T = W_{r}+W_{b}+W_{d}+W_{a}+W_{p},
    \label{weight composition}
\end{equation}
where $W_{r}$, $W_{b}$, $W_{d}$, $W_{a}$, and $W_{p}$ denote the weights of the robot frame, battery, deforming mechanical module, avionics devices, and payload, respectively. In some cases, increasing battery capacity can improve flight duration as more energy is provided. Flight time, however, may decrease if the battery is too heavy. The relationships between voltage, instantaneous current, and thrust (acquired by curve fitting) were used to develop a mapping relationship between flight duration and battery weight, as shown in Fig. \ref{flight duration modeling}(c). 

\subsection{Power Setting}

In this study, we strictly confined the robot as light as possible, and used two frequently-used lightweight optimization methods: light material selection and computer-aided structural optimization \cite{zhu2018light}. Finally, the robot's weight is about 17 kg, excluding the battery. The results showed a robot with an MN805S-170KV power train and a $\sim$5.7 kg battery achieved the longest hovering duration ($\sim$21 minutes). Detailed relationships between the total robot weight and the battery weight utilizing T-Motor MN805S motors are provided in Fig. \ref{duration weight relationships}. The no-fly zone represents areas in which the robot cannot fly in a given configuration, due to excessive weight. In contrast, the lines of varying duration show weight configurations that can achieve flight. The proposed design is indicated by the red dot in the figure.

\section{Deformable Structure Design}

Linkages\cite{Hu2018a} and putters\cite{machines10050289} have successfully been used to actuate deformable structures in previous studies. However, redundant components create additional weight and can lead to more failures. As such, the proposed robot uses servos to drive the arms directly. Meanwhile, locking mechanisms and locking propeller motors are also highlights of the novel deformable structure. 

\subsection{Foldable Arms}
A propeller's efficiency is significantly influenced by its diameter, as more power and thrust can be produced by larger propellers. For this reason, our robot uses 26-inch propellers to generate more thrust. However, propellers that are too large can cause issues for ground mobility, making it difficult to maneuver in confined spaces. In addition, a wide mass distribution can reduce the dynamic stability of the robot and induce vibrations \cite{saab2018robotic}. As such, we propose a new type of oblique folding design in which the front arm is folded downward at an angle of 20 degrees and the rear arm is positioned horizontally to avoid physical interference between the front and rear arms. As the four arms rotate 135 degrees outward on the folding servo, the flight module unfolds and the platform enters flight mode, forming a square standard quad-rotor configuration.

\begin{figure}[t]
    \centering
    \includegraphics[width=7cm]{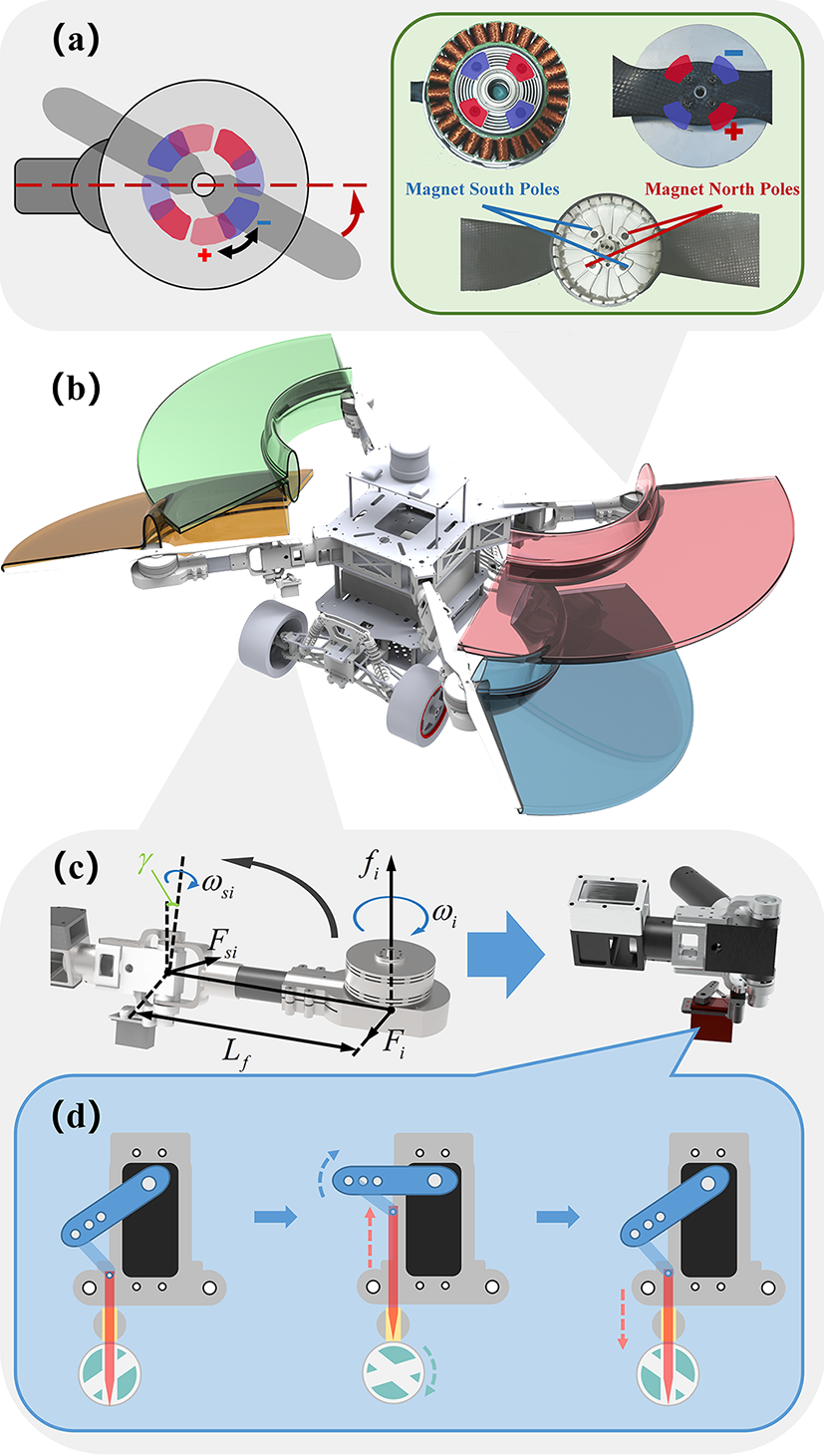}
    \caption{A demonstration of the automated deformable structure. (a) A fixed propeller motor. (b) Motion paths in 3D space for the deformable structures. (c) A display of the deformable forearm. (d) The design and function of the locking mechanism.}
    \label{deformable structure}
         \vspace{-5 mm} 
\end{figure}

Fig. \ref{deformable structure}(b) shows the deformable spatial motion path for the robot arm. The four arms form a circular path around the rotation axis of the deformable mechanism. Unlike in previous studies \cite{Hu2018a}, proposed robot uses a unique motor for the fixed propeller, shown in Fig. \ref{deformable structure}(a) and Section III.C, to maintain the relative propeller angle in the same direction as the arm.

Fig. \ref{deformable structure}(c) shows the deformable mechanisms of the robot arms. As the motors begin to rotate, they generate tremendous amounts of torque at the dual-axis steering gears on the arms. The Newton-Euler equation suggests a dynamic model for the quad-rotor aircraft can be expressed as:
\begin{equation}
M_{b}=J_{b} \dot{\omega}_{b}+\omega_{b} \times J_{b} \omega_{b}+M_{g}+M_{d}.
\label{quadrotor}
\end{equation}
The synthesis of gyroscopic torque $M_{g}$ and aerodynamic friction torque $M_{d}$ can then be represented by:
\begin{equation}
M_{g}=\sum_{i=1}^{4} \overrightarrow{\omega_{b}} \times J_{b}\left[0 \quad 0(-1)^{i+1} \Omega_{i}\right]^{T},
\label{synthesis of gyroscopic effect torque}
\end{equation}

    \vspace{-3 mm}
    
\begin{equation}
M_{d}=\operatorname{diag}\left(d_{\phi}, d_{\theta}, d_{\psi}\right) \dot{\zeta},
\label{aerodynamic friction torque}
\end{equation}
where $J_{b}=\operatorname{diag}\left(d_{\phi}, d_{\theta}, d_{\psi}\right)$ is the moment of inertia for each rotor and $d_{\phi}$, $d_{\theta}$, and $d_{\psi}$ are the corresponding aerodynamic damping coefficients.

Complex deformable arms can affect the payload and reliability of land-air robots as each rotor is fixed to the main body of the robot by an independent arm. The propeller will then generate yaw forces during rotation, relying on the force generated by the steering gear to maintain the position of arms. Equations (\ref{quadrotor})-(\ref{aerodynamic friction torque}) suggest the moment of inertia for the arm can be expressed as:
\begin{equation}
\begin{array}{l}
J_{f i}=\frac{m_{i}}{12} L_{f i}^{2} \times \cos (\gamma),
\end{array}
\end{equation}
    \vspace{-4 mm}
\begin{equation}
\begin{array}{l}
J_{r i}=\frac{m_{i}}{12} L_{r i}^{2},
\end{array}
\end{equation}
where $m_{i}$ is the mass of the arm and $J_{i}$ is the moment of inertia for the arm around the axis of the steering gear. $L_{fi}$ and $L_{ri}$ indicate the length of the front arms and the rear arms, respectively. The dynamic equation for the deformable arm is given by:
     \vspace{-1 mm}
\begin{equation}
\operatorname{diag}\left(m_{i}\right)\left[\begin{array}{c}
\ddot{x} \\
\ddot{y} \\
\ddot{z}
\end{array}\right]=\left[\begin{array}{c}
F_{x i} \\
F_{y i} \\
F_{z i}
\end{array}\right],
\end{equation}
where $F_{x i}, F_{y i}, \text { and } F_{z i}(i=1 \ldots 4)$ are the decoupling forces along the x-axis, the y-axis, and the z-axis at the rotating shaft of each motor, respectively. The rotation matrix decouples these torques.

The reference coordinate is a global coordinate system. When the arm and the rotor rotate relative to the fuselage coordinate system, the inertia tensor for the arm will not rotate. This tensor can be expressed as:
\begin{equation}
J_{a r m, i}=R_{z}\left(\theta_{i}\right) J_{a r m} R_{z}\left(\theta_{i}\right)^{T},
\end{equation}
where $R_{z}$ is the rotation matrix around the z-axis of the global coordinate system. The yaw angle is $(\theta_{i}), i=(fl,fr,rl,rr)$, and $fl, fr, rl, rr$ demonstrate front left arm, front rear arm, rear left arm, and rear right arm, respectively. The moment of inertia for the motor does not change as it rotates around its z-axis.

The dynamic model of the folding mechanism is the base of the simulation experiments in Section V. Meanwhile, it supports the robot in calculating and monitor the actuator's real-time operating performance, detecting performance degradation as early as possible, and enabling predictive maintenance.

\subsection{Locking Mechanism}
A physical locking mechanism was included in he design to improve stability and ensure the safety of the robot. As shown in Fig. \ref{deformable structure}(d), the machine maintains 2 lock holes at key positions on the arms. The steering gear is then fixed on the rack and drives the auto-locking mechanism, while controlling pin movements by modifying draw bar rotation. These lock holes were denoted limiter-1 and limiter-2. Limiter-1 prevented the pin from deviating off the motion tracking course. Limiter-2 contained two key position holes, corresponding to the expanding and folding modes of the deformable arm. This pin could be accurately inserted into the holes by rotating the arm. This structure increased the operating safety of the platform in flight mode and effectively reduced the working pressure on the dual-axis steering gear.

\subsection{Fixed Propeller Motor}

Unsecured propellers can cause collisions with the robot body during deformations. To address this issue, we designed a motor to ensure the propeller blades remained oriented in the direction of the arms. The motor included 2 south and 2 north magnetic poles added to the inner and outer rotors. As shown in Fig. \ref{deformable structure}(a), the propellers were fixed to the outer rotors, which relied on the magnets to quickly reset the propeller and keep it parallel to the arm at all times. This design prevented excessive energy loss during propeller rotation.

\section{Overall Motion and Control Strategy}

\begin{figure}[h]
    \centering
    \includegraphics[width=8.6cm]{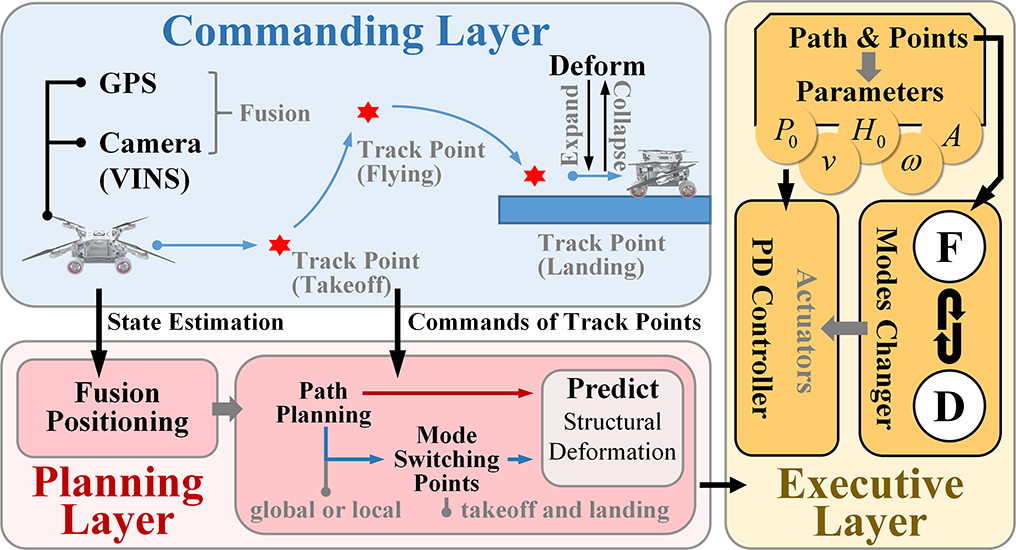}
    \caption{The motion and control strategy is divided into three layer: commanding layer, planning layer, and executive layer.}
    \label{motion}
\end{figure}

Fig.\ref{motion} shows the robot's overall motion and control strategy, including the mode switching methods. Commanding layer provides the abstract commands from the operators or ground station. At present, the robot can only rely on the operator's commands to perform the tasks of mode switching in a relatively open environment. When the platform is optimized in the future and possesses sufficient perception and planning capabilities, it can performs more autonomous movements. The operator provides the end point or track points to lead the robot's motion and as input for motion control and planning tasks.

The planning layer relies on the fusion positioning of GPS and visual odometer (VINS\cite{qin2018vins}) to determine its relative three-dimensional space information and whether it is on the ground. Lidar is used to perceive the surrounding three-dimensional environment information to assist navigation and obstacle avoidance tasks. The positioning and track points are then used to generate the motive path and critical points, including takeoff and landing points. According to the motion state, the robot predicted the structural deformation time in advance at the mode switching point or during the motion process. Arms can be expanded or collapsed at mode switching points or during travel.

The path and track points then put into the executive layer and convert to the mathematical parameters with robot's states and expection, including horizontal spatial position $P_0$, height $H_0$, attitude $A$, velocity $v$, and angular velocity $omega$. The parameters are input by the modes changer and controller to switch the motive mode between the flying mode and the driving mode, meanwhile driving the actuators.

The strategies structure of the three-layer can be mapped to the robot's mechatronics system. The sensors and communications models refer to the commanding layer, the edge computing device and the planning algorithm refer to the planning layer, and the controller and actuators refer to the executive layer. The combined strategy supports the robot can perform autonomous ground and air movement and mode switching.

\section{Experimental Evaluation}

The robustness and stability of the automated foldable arms were evaluated by simulations and experiments as part of the study. The resulting flexibility for land-air composite tasks was assessed using a 3D A$^\ast$ search algorithm.

\subsection{Deformable Structure Experiments} 

This subsection describes the simulated validation of the proposed deformable structure. We measured and identified the mass, material, spatial distributions and others parameters in Tab. \ref{specification} for each structure. These data were used to construct a detailed co-simulation using ADAMS and MATLAB. 

\subsubsection{Motion simulation}

\begin{figure*}[t]
\centering
\includegraphics[width=16cm]{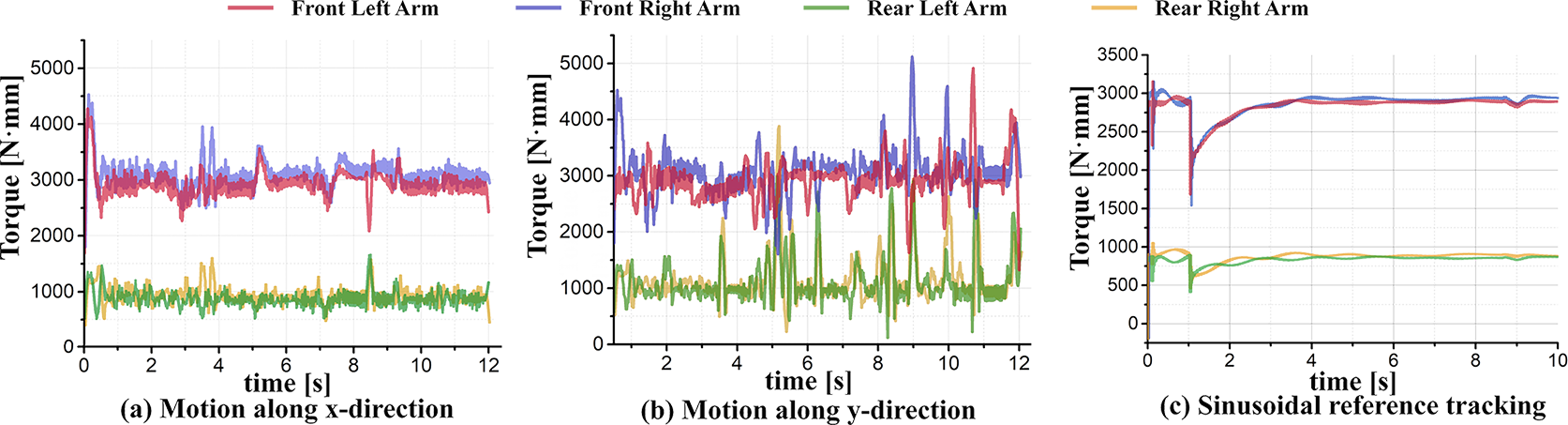}
\caption{Simulated torque on the deformable robot structure. (a) Accelerated motion in the x-direction. (b) Accelerated motion in the y-direction. (c) The robot performing a wave-like motion.}
\label{motion_simulation}
    \vspace{-4 mm}
\end{figure*}

\begin{figure}[h]
    \centering
    \includegraphics[width=6cm]{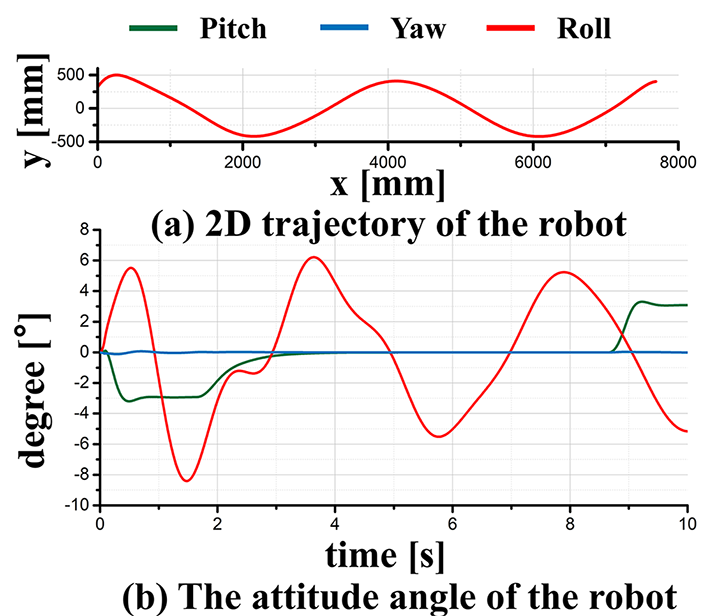}
    \caption{Robot motion during sinusoidal reference tracking. (a) A top-view of the 2D robot trajectory. (b) The robot's attitude angle.}
    \label{sin}
\end{figure}

Defined the front direction of the initial state of the robot as the two coordinates' x-direction, and established the global and robot coordinate systems with the right-hand rule. The robot then conducted three sets of experiments to obtain torques for the deformable arm structure. Figs. \ref{motion_simulation}(a) and (b) show the deformable servos' torque produced during motion along the x and y axes. In both sets of experiments, the robot was tilted 15 $degrees$ towards the x-axes and y-axes and moved along the directions. Fig. \ref{motion_simulation}(c) shows torque variations while tracking a sinusoidal path, with trajectory and attitude changes shown in Fig. \ref{sin}. Torques generated on the front arms were primarily concentrated at 3000 $N \cdot mm$ and torques on the rear arms were near 1000 $N \cdot mm$. The torque on the deformable structure was also much smaller than the upper limit for the servos.

\subsubsection{Thrust simulation}

In the experiments, motor rotary speed was gradually increased from a standstill, reaching a maximum speed of 8000 $rad/min$. The torque generated on the deformable mechanism is a critical parameter for robot-safe operation and detection. A torque of 1000 $N \cdot mm$ was generated at rest, due to the inclination of the front arm. An increase in motor speed produced maximum torque on the front and rear arms of 6706 $N \cdot mm$ and 1711 $N \cdot mm$, respectively, far from the upper limits of the design 7800 $N \cdot mm$.

\subsection{Multi-modal Locomotion Planning}

\begin{figure}[t]
    \centering
    \includegraphics[width=8.5cm]{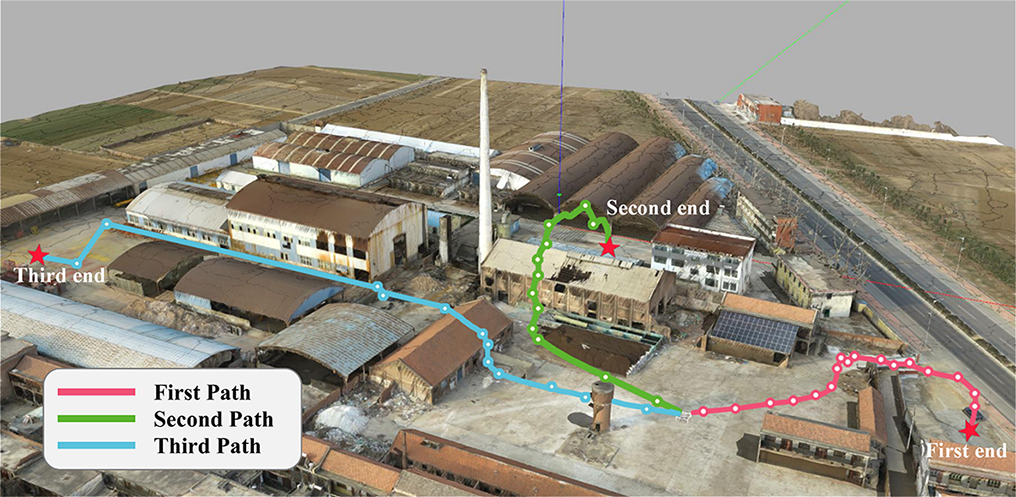}
    \caption{The experiments demonstrate and verify the superiority and flexibility of the multi-modal robot in complex scenes.}
    \label{planning}
     \vspace{-6 mm} 
\end{figure}

The flexibility of the robot was evaluated using a rescue scenario in an abandoned factory, which required the robot to search an optimal hybrid path to reach a specified target point. A DJI phantom 4 Pro was used to map the target area. The map was then rasterized, the obstacles were inflated, and the raw planned path was smoothed to satisfy the dynamic limits of flying and driving. As shown in Fig. \ref{planning}, the robot originated from the center of the factory and attempted to reach three end points along the generation paths. The robot utilized multiple motion modes to fly over obstacles yet conserve energy, since the destinations were relatively isolated from the starting point. A simple rule-based algorithm was implemented based on 2D/3D A-star path planning \cite{duchovn2014path,yang2014literature}. Energy consumption was considered in the process of heuristic algorithm design, so the robot would tend to drive on the ground rather than flying. As such, the robot demonstrated its adaptability and flexibility to various environments.

\begin{figure*}[t]
\centering
\includegraphics[width=17cm]{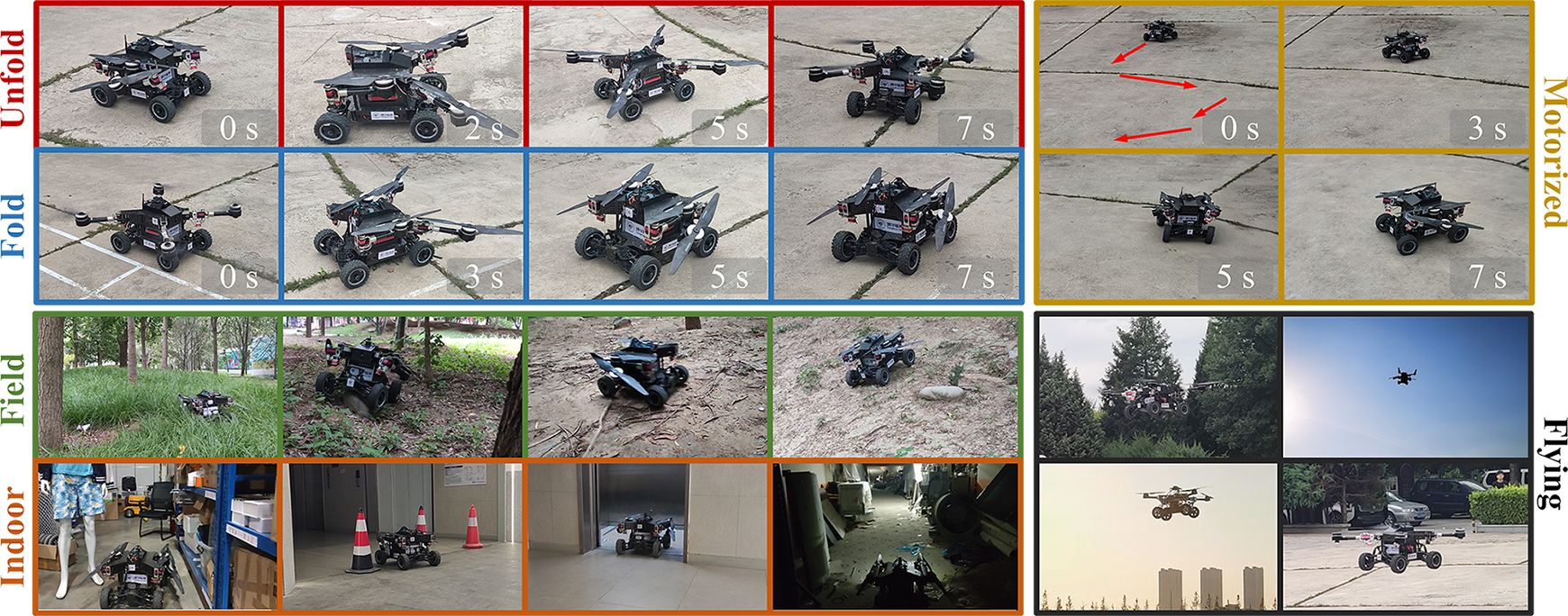}%
\caption{Robot platform experiments.}
\label{experiments}
\vspace{-4 mm}
\end{figure*}

\subsection{Platform Tests}

Experiments were conducted in both field and indoor environments under various scenarios, to verify performance. The corresponding results are shown in Fig. \ref{experiments} and a video is included in the link in Section I. 

\subsubsection{Mode transition}
The first set of tests illustrated the transition process between flying and driving modes (the red and blue frames in Fig. \ref{experiments}). The robot only required 5 seconds to complete the deformation process. This practical experiment showed the robot could rapidly fold the arms and switch motion modes. The propellers could also be well positioned to avoid interference during transition.

\subsubsection{Driving in different scenarios}
In this experiment, the robot performed ground motion tests on flat roads, in unstructured scenes in the field, and in complex indoor environments. The robot exhibited a significant advantage in moving speed on structured paved roads, with a maximum speed reaching over 10 $m/s$. Fig. \ref{experiments} (see the yellow frame in Fig. \ref{experiments}) demonstrates highly flexible maneuvering on ground. In addition, the robot was tested on loose roads and in indoor human activity scenarios. It could easily pass through grasses, woods, and sandy ground, crossing a slope of more than 30° without slipping. However, it is restricted by its tire height, which made it challenging to overcome obstacles more than 100 mm in height (see the green frame in Fig. \ref{experiments}). The robot was also tested in warehouses, elevators, and dark corridors, in which it performed reliably and stably (see the orange frame in Fig. \ref{experiments}).

\subsubsection{Flying tests}
Flight tests were conducted outdoors, including takeoff, landing, acceleration, emergency braking, and path tracking (see the black frame in Fig. \ref{experiments}). During movement, the robot always maintained a normal posture and the desired control. In addition, the deformation mechanism operated stably without triggering an intervention from the locking mechanism. Some shaking did occur during takeoff, due to ground spoilers. Meanwhile, we measured and estimated the robot's average power consumption during hovering and driving. The result shows that the power consumption of driving (2840 $W$) is 54.75 \% lower than that of flying (6276 $W$).

\begin{figure*}[b]
\centering
\includegraphics[width=18cm]{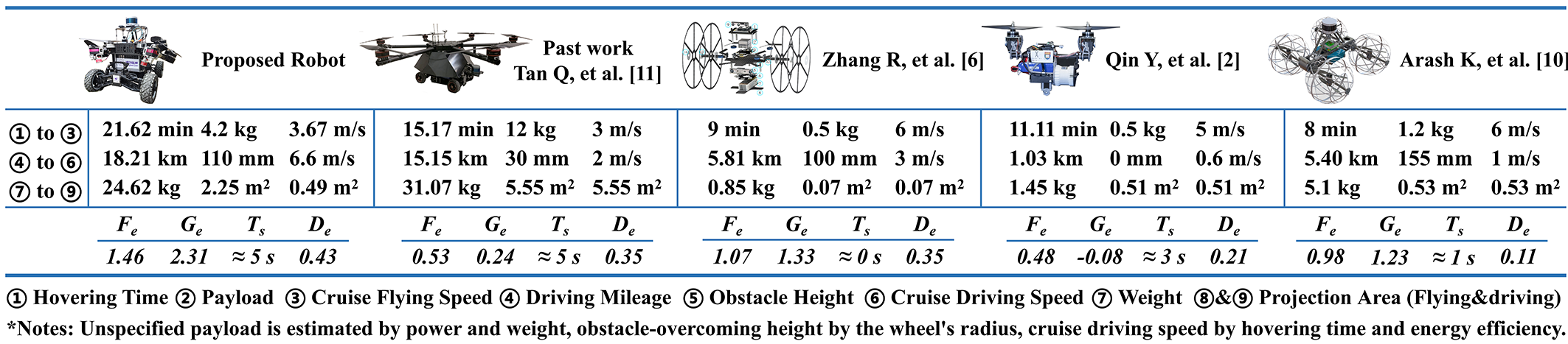}%
    \caption{Comprehensive performance quantitative evaluation of proposed robot, past work and typical land-air robot.}
\label{comparision}
\vspace{-4 mm}
\end{figure*}

\subsection{Evaluation and Comparison}

In this subsection, we compare the proposed robot with other related works. The robot's size, duration, payload, and movement abilities are all critical manifestations of its performance \cite{nie2013robots}. Meanwhile, flight and ground movement capabilities need to be comprehensively evaluated. As such, we removed unit inconsistencies between all parameters and evaluated proposed robot with following indexes to obtain the comprehensive performance of the land-air robot \cite{tan2021multimodal,zhang2022autonomous,qin2020hybrid,kalantari2020drivocopter}.

\subsubsection{Comprehensive flight ability}
Hovering flight duration $t_f$, payload $p$, and maximum tracking speed (air) $v_{f}$ are critical indicators of the robot's index of flight capability $F_e$. They are also positively correlated with the robot's vertical projection area of flight state. $w_{i}$ indicates the weight of the corresponding indicator set according to the task requirements, and this paper uses the same weight as $1$. We use Min-Max normalization method to evaluate the comprehensive flight capability as follow

\begin{equation}
n(x)=\frac{x_{i}-x_{\min }}{x_{\max }-x_{\min }}
\end{equation}

where $x_i(i= 1 \cdots n)$ express the parameters of robot $i$. $x_{max}$ and $x_{min}$ are the max and min values of the robots' same parameters.

     \vspace{-3 mm} 
     
\begin{equation}
F_{e}\left(t_{f}, p, v_{f}, S_{l a n d}\right)=\sum_{i=t_{f}, p, v_{f}} w_{i} \cdot n(i)-w_{s} \cdot n\left(S_{\text {land }}\right)
\end{equation}

\subsubsection{Comprehensive driving ability}
Similarly, the index of comprehensive ground performance $G_e$ of the robot is evaluated by the ground driving mileage $m_d$, the obstacle-surmountable height $h_o$, and the maximum tracking speed (land) $v_{d}$.

     \vspace{-4 mm} 
     
\begin{equation}
G_{e}\left(m_{d}, h_{i}, v_{d}, S_{\text {air }}\right)=\sum_{i=m_{d}, h_{i}, v_{d}} w_{i} \cdot n(i)-w_{s} \cdot n\left(S_{\text {air }}\right)
\end{equation}

\subsubsection{Modes switching time}
Modes switching time $T_s$ express the time that robot switches from driving mode to flying mode or from flying mode to driving.

\subsubsection{Comprehensive duration}
Comprehensive duration in flying mode $t_f$ and driving mode $m_d$ are crucial for land-air robots' ability to undertake more productive tasks. And the duration is also strongly related to robots' weight $G$. Therefore, the land and air duration time $D_e$ is comprehensively considered and used as a single evaluation index.

\begin{equation}
D_{e}\left(m_{d}, t_{f}\right)=\sum_{i=m_{d}, t_{f}} w_{i} \cdot n(i)-w_{G} \cdot n(G)
\end{equation}

The results (Fig. \ref{comparision}) show that our robot got higher scores on $F_e$ and $G_e$, thus that it has relatively comprehensive land and air capabilities. Meanwhile, the robot also got a slightly higher score on $D_e$. Although the mode switching speed of the robot is slow and the energy efficiency is slightly below than \cite{zhang2022autonomous,qin2020hybrid}, our design conception improve the robot's comprehensive abilities obviously. It should be noted that the comparison here is only a reference for the robot's capabilities. The parameters' weights in the method also need to be adjusted according to the specific task requirements. For example, \cite{zhang2022autonomous} can perform more efficient and safe exploration in narrow and complex scenarios. When evaluating such robots, we can increase the weights of flight and size. The evaluative method can also introduce quantitative indicators such as climbing performance and security.

\section{Conclusion}

This study proposed a multi-modal land-air robot with automatic deformable arms for the first time, with the design conception which solving the compatibility issues for the hybrid land-air robot. By estimated and calculated the powers and energy, the robot got a suitable battery and power settings. Meanwhile, a simple yet reliable foldable structure was used to significantly expand the space for a flight mechanism and increase the thrust. This approach also reduced the size of the driving mode and allowed the platform to traverse narrow corridors. The robot was tested in cement pavement, field and indoor, we knew that the driving consumption is 54.75 \% lower than flying consumption. Compared with other land-air robots by proposed evaluation methods, our robot offers better comprehensive for flying ability, as well as improved roadway driving capabilities. The robot had a good balance of flight and driving comprehensive capabilities. Furthermore, the mechatronics design concept of the robot can help other land and air robots improve their comprehensive motion capabilities.

\bibliographystyle{IEEEtran}
\bibliography{0article}

\end{document}